 \documentclass[10pt,usletter]{article}
 \usepackage[totalwidth=420pt,totalheight=625pt]{geometry}

\newcommand{\cv}[1]{}
\newcommand{\av}[1]{#1}

\av{\usepackage{amssymb}}
\usepackage{amsmath}
\usepackage{booktabs}
\av{\usepackage{url,hyperref}}
\usepackage{listings}
\av{\usepackage{xcolor}}
\usepackage{tikz} %
\av{\usepackage{enumitem}}
\usetikzlibrary{shapes.geometric, arrows.meta, positioning}

\av{
 \bibliographystyle{plainurl}
 \usepackage[numbers,sort&compress]{natbib}
 \usepackage{comment}
 \excludecomment{CCSXML}
}

\title{CP-Agent: Agentic Constraint Programming\thanks{To appear in
    the proceedings of the International Workshop on Agentic
    Engineering (AGENT@ICSE 2026).
Supported by the Austrian Science Fund (FWF) projects 10.55776/P36420 and 10.55776/COE12.
}}

 \author{Stefan Szeider\\[4pt]
  \small Algorithms and Complexity Group\\[-3pt]
  \small TU Wien, Vienna, Austria\\[-3pt]
  \small \href{https://www.ac.tuwien.ac.at/people/szeider/}{www.ac.tuwien.ac.at/people/szeider/}
 }
 \date{}

\begin{document}

\av{\maketitle}

\begin{abstract}
 The translation of natural language to formal constraint models
 requires expertise in the problem domain and modeling frameworks. To
 explore the effectiveness of agentic workflows, we propose CP-Agent,
 a Python coding agent that uses the ReAct framework with a
 persistent IPython kernel. We provide the relevant domain knowledge
 as a project prompt of under 50 lines. The algorithm works by
 iteratively executing code, observing the solver's feedback, and
 refining constraint models based on execution results.

 We evaluate CP-Agent on 101 constraint programming problems from
 CP-Bench. We made minor changes to the benchmark to address
 systematic ambiguities in the problem specifications and errors in
 the ground-truth models. On the clarified benchmark, CP-Agent
 achieves perfect accuracy on all 101 problems. Our experiments show
 that minimal guidance outperforms detailed procedural
 scaffolding. Our experiments also show that explicit task management
 tools can have both positive and negative effects on focused
 modeling tasks.
\end{abstract}

\cv{\maketitle}

\section{Introduction}\av{\thispagestyle{empty}}

Constraint Programming is a declarative paradigm for solving
combinatorial problems in which users specify the constraints a solution
must satisfy, rather than prescribing how to find
it~\cite{Wallace1996}. The main obstacle preventing wider adoption of
CP remains the modeling bottleneck. Converting natural language
problem descriptions into formal constraint models requires deep
expertise in both the problem domain and the modeling
framework~\cite{Freuder2014}. This challenge is referred to as the
``holy grail'' of constraint programming~\cite{Tsouros2023HolyGrail}.

In recent years, Large Language Models (LLMs) have been successfully
applied to automate this translation process. We considered
CP-Bench~\cite{Michailidis2025}, a benchmark introduced to evaluate
LLM-driven constraint modeling capabilities, which contains 101
diverse combinatorial problems sourced from various established CP
community resources. These problems are used as benchmarks for
optimization and satisfaction tasks, and include a variety of distinct
constraint types. This benchmark addresses the limitations of existing
datasets, such as NL4Opt~\cite{Ramamonjison2022} and Logic Grid
Puzzles~\cite{Tyagi2024}, which lack the diversity and complexity
needed to evaluate constraint modeling capabilities.

Michailidis et al.~\cite{Michailidis2025} evaluated several LLM-based
workflows on CP-Bench. With direct prompting, Python-based frameworks
(CPMpy, OR-Tools) achieve overall good accuracy up to 65\%. Indeed,
MiniZinc (a popular constraint modeling language) peaked at
around~50\%. Repeated sampling can be combined with self-verification
to improve accuracy for Python-based frameworks to approximately~70\%.

The relationship between fixed and agentic workflows is now well
understood. Chain-of-thought prompting~\cite{Wei2022} and subsequent
work~\cite{Yao2023,Schick2023} have been widely used in the
literature. One distinguishes between \emph{fixed} workflows that use
well-scoped, short-horizon tasks and \emph{agentic} workflows where
the LLM runs in a loop and uses tools. We hypothesize that constraint
modeling from natural language falls into the category of tasks that
require an agentic workflow with iterative refinement, hypothesis
testing, and dynamic adaptation based on solver feedback. Our
experiments show that this is indeed the case.

To test this hypothesis, we introduce \emph{CP-Agent}, a
constraint modeling agent that utilizes our general-purpose
\emph{agentic-python-coder} system. CP-Agent
maintains a persistent \textit{IPython} kernel~\cite{Perez2007},
enabling stateful and iterative development. Instead of
domain-specific logic embedded in its architecture, we use a concise
project prompt as the guide. The agent can execute code, receive runtime feedback
from the constraint solver, and adjust its constraint model accordingly.

The underlying coding agent is available as an \emph{MCP
 server}~\cite{MCP2024}. The Model Context Protocol (MCP), now
governed by the Linux Foundation, is widely used for connecting
language models with external tools~\cite{Hou2025MCP}. With this
integration, the CP agent can be used from any MCP-compatible client,
including coding assistants like Cursor or AI chatbots like Claude
Desktop.

On the CP-Bench dataset of 101 problems (with added clarifications for
some ambiguous problems and output format requirements), CP-Agent
solves all problems correctly. We use Claude Sonnet 4.5 as the primary
reference language model, though we also tested this with other
models from different providers. Just by changing the prompt while
keeping the agent architecture, we can adapt the system to different
domains. This is exemplified by follow-up
work~\cite{Szeider2025ASPBench} on modeling tasks for logic
programming.

In addition, we provide an empirical evaluation of using a task
management tool within our agentic coding workflow. The results are
mixed: task management is helpful for some but counterproductive for
others; the net effect was adverse, with significantly
increased token usage, suggesting that the overhead can distract as
often as it organizes.

\section{Related Work}

Our approach follows the neurosymbolic paradigm, combining neural AI
with symbolic constraint satisfaction~\cite{Regin2024CPLLM}. The
LLM-based agent converts natural language into formal constraints. The
CP solver guarantees formal correctness of the computation.

LLM agents have been explored for software engineering
tasks. SWE-agent~\cite{DBLP:conf/nips/YangJWLYNP24} introduced the
Agent-Computer Interface, a file-system-based approach for resolving
GitHub issues that allowed navigating and editing repository
files. Our approach is based on single-script generation using a
stateful IPython kernel, which eliminates file I/O overhead for this
class of problems.
CodeAct~\cite{DBLP:conf/icml/WangCYZLPJ24} promotes executable Python
code as a unified action space, and finds that this approach still
outperforms more restrictive formats. We apply this general approach
to constraint modeling, where a general action space can solve
specialized tasks using focused prompts.
The iterative refinement that we use in CP-Agent is related to
automated program repair. ChatRepair~\cite{DBLP:conf/issta/XiaZ24} proposed conversational repair, where one iteratively refines patches based on test failures.
RepairAgent~\cite{DBLP:conf/icse/BouzeniaDP25} considers repair as an autonomous task with tools for code search, patching, and testing. Our approach is similar in spirit to iterative code repair, where refinement is used to correct errors. Still, we apply it to generative tasks, translating natural language into constraint programs rather than repairing existing code.

The validity of an evaluation depends on benchmark
quality. EvalPlus~\cite{DBLP:conf/nips/LiuXWZ23} showed that popular
benchmarks have insufficient tests, while augmented test suites reveal
previously undetected errors. Only recently,
ClassEval~\cite{ICSE24:ClassEval} proposed a new benchmark for
structural complexity through manual curation.
We address problem ambiguity in the original CP-Bench problems and
propose a fixed version of the benchmark that allows more rigorous
testing of constraint modeling capabilities.

\section{Agentic Python Coder}

Our CP-Agent is based on a generic Python coding agent, equipped with
a special prompt for constraint modeling. Our coding agent
implements the basic idea of ReAct~\cite{Yao2023} (Reason and Act) in a way that is both minimal and capable. The architecture can be adapted to new problem domains through prompt engineering rather than code modification. By this separation, domain-agnostic agent capabilities remain distinct from task-specific knowledge.
Our implementation is publicly available on GitHub and via PyPI for installation via ``\texttt{pip install agentic-python-coder}''~\cite{AgenticCoder2025}.

\subsection{Architecture}

\paragraph{Virtual execution space} Unlike software engineering
agents, which navigate complex file trees, we do not maintain multiple
files; instead, development is focused on a single file. We process
development, refinement, and execution in a virtual space provided by
the IPython kernel~\cite{Perez2007}, which runs as a subprocess. The
kernel provides a persistent state across multiple code
executions. This guarantees that variables, functions, and imports
remain accessible throughout a session and enables rapid hypothesis
testing without file I/O overhead. The agent proceeds in rounds, and
each round consists of code manipulation followed by immediate
feedback via the Read--Eval--Print Loop (REPL). This design is well-suited for developing constraint programming models and similar
focused generation tasks.

The ReAct agent is implemented using LangGraph modules. LangGraph provides an important link between workflow
management and durable execution, enabling the development of stateful
agents. This even prevails when we consider the ReAct
pattern~\cite{Yao2023} as interleaved reasoning and acting, which
allows the agent to interact with the execution environment and adjust
solutions through iterative cycles.

We will now explain the details of each component: the
\emph{ReAct} loop controller, which implements the reasoning-action
cycle (using LangGraph~\cite{LangGraph2024}); three
\emph{tools} the agent can call; and a \emph{project prompt} that
provides domain-specific guidance. In each iteration of the ReAct
loop, the agent reasons about the task, selects and executes tools, and
observes results until a final solution is obtained and saved to a
file (Figure~\ref{fig:react}).

\begin{figure}[htb]
\centering
\tikzstyle{startstop} = [ellipse, minimum width=2cm, minimum height=0.5cm, text centered, draw=black!70, fill=blue!10]
\tikzstyle{process} = [rectangle, minimum width=2cm, minimum height=0.5cm, text centered, draw=black!70, fill=blue!10, rounded corners=3pt]
\tikzstyle{endnode} = [ellipse, minimum width=2cm, minimum height=0.5cm, text centered, draw=black!70, fill=blue!30]
\tikzstyle{arrow} = [thick,->,>=stealth,black!70]
\tikzstyle{dashedarrow} = [thick,->,>=stealth,black!50,dashed]
\begin{tikzpicture}[scale=0.80]
\node (start) [startstop] {Task};
\node (llm) [process, below of=start] {Reason};
\node (tools) [process, below left of=llm, xshift=-1.5cm, yshift=-0.5cm] {Act (Tools)};
\node (end) [endnode, below right of=llm, xshift=1.5cm, yshift=-0.5cm] {Solution};
\draw [arrow] (start) -- (llm);
\draw [arrow] (llm) to[bend left=20] node[anchor=west, pos=0.5, font=\small] {execute} (tools);
\draw [arrow] (llm) to node[anchor=west, pos=0.4, font=\small] {~~complete} (end);
\draw [arrow] (tools.north) to[bend left=30] node[anchor=east, pos=0.5, font=\small] {observe~~} (llm.west);
\end{tikzpicture}
\caption{The ReAct (Reason and Act) framework employed by the agent.}
\label{fig:react}
\end{figure}

\paragraph{Tools}
The main tool is \texttt{python\_exec(code:~str)}, which runs
Python code in the persistent kernel environment and returns
stdout/stderr for solver feedback. Each call to this tool builds
incrementally upon the previous calls. The function
\texttt{save\_code} transforms the solution into a file on disk,
signaling completion. One could omit the \texttt{save\_code} tool
since file access is also possible via
\texttt{python\_exec(code:~str)}. However, having a dedicated tool
for code saving has the advantage of easier file-path handling
and makes it clear to the agent that calling this tool marks the end
of its session. Optionally, we give the agent access to a task
management tool to keep track of tasks in the form of a ``to-do list.''
The tool call has the format \texttt{todo\_write(todos:~List[Dict[str,
 Any]])}, where todos is the task list, and each entry specifies task
attributes. By construction, each task must contain \texttt{id},
\texttt{content}, \texttt{status}, and \texttt{priority} fields, and
only one task can be marked as ``in\_progress'' at any time.

\paragraph{Dynamic dependency injection} We implement a custom kernel
manager, which wraps the standard IPython kernel launch. Instead of
having the agent run pip install commands or rely on prebuilt container images, we inject packages at startup. Using the
\texttt{uv} package manager~\cite{uv2025}, we split package
installation into an ephemeral environment that runs at kernel startup. The
kernel resolves dependencies based on session configuration; each
session runs in a clean environment with only the specified packages.

\paragraph{LLM integration}

The LLM that drives the agent is provided via the
OpenRouter API. This allows us
to use models from different model-makers through a uniform API and
configuration, and with a single API key. We record all the agent's steps
and tool calls in log files. These logs provide valuable insight into
the agent's problem-solving process and resource
consumption.

\paragraph{Prompt architecture}
The agent operates with the following three-fold prompt hierarchy: the
\emph{system prompt} (approximately 100 lines) specifies how the agent
can be used to reason about code and execute iterative development
cycles, defining general coding behavior and tool usage patterns
without task-specific specialization. The \emph{project prompt}
combines domain-specific knowledge (such as constraint modeling) with
examples and best practices relevant to this particular problem
class. The concrete task to solve is in the \emph{task prompt},
typically read from a \texttt{task.md} file in the working directory
or provided directly via a command line parameter.
The same agent
infrastructure can tackle diverse problem domains by simply replacing
the project prompt.

\section{CP-Agent}

For constraint programming, the generic coding agent is equipped with a domain-specific project prompt that instructs it to use CPMpy as the modeling framework. No specialized constraint programming
logic is embedded in the architecture. The agent can create and test
simplified problem instances and verify the constraint model on
example data. The agent can even implement custom validation logic by
refining models based on solver feedback. This flexible workflow,
together with the persistent execution environment, provides an
excellent environment for discovering and correcting modeling errors
that would escape detection in single-pass generation approaches.

\subsection{CPMpy}

We use CPMpy~\cite{Guns2019} as the constraint programming foundation
primarily due to its Python integration. This library is based on
NumPy for efficient handling of decision variables as arrays, making it well-suited for integration with the broader Python ecosystem. Its
solver-agnostic design allows a more refined selection of backends and
hence enables modeling problems once and solving them using various
solvers, including CP, MIP, SMT, and SAT. This design is in line with
findings from Michailidis et al.~\cite{Michailidis2025} that LLMs
perform better with Python-based frameworks than with domain-specific languages, likely due to the billions of lines of Python code in their training data.

\subsection{Project Prompt}

\sloppypar The project prompt (\texttt{cpmpy.md}), under 50 lines, provides
succinct guidance for constraint modeling. The complete prompt is
available online at the GitHub repository~\cite{AgenticCoder2025}.
The prompt contains the following core rules: use CPMpy's declarative
modeling rather than implementing search algorithms, output only valid
JSON using \texttt{json.dumps()}, always import the json Python module
when outputting JSON, check the exact required output format, and manually test solutions to ensure they satisfy the problem requirements. A basic code template demonstrates the standard structure, and brief sections cover three essential CPMpy constraints (\texttt{AllDifferent}, \texttt{sum}, \texttt{Circuit}) and optimization via \texttt{model.minimize()} and \texttt{model.maximize()}.

This minimal prompt is based on an ablation study. We initially
experimented with a substantially longer prompt of approximately 800
lines, containing detailed modeling procedures, extensive problem-type
catalogs, and debugging guidance. The shorter version performs on par with, or better than, the longer version across all performance metrics. This
indicates that for modern LLMs with constraint-programming knowledge
in their training data, concise guidance is sufficient; excessive
detail may distract and consume input tokens without providing benefit.

\section{Benchmark Problems}

The CP-Bench collection~\cite{Michailidis2025} provides an excellent
benchmark for testing automated constraint modeling capabilities. This
collection comprises 101 constraint programming problems from four
established sources: CSPLib~\cite{Gent1999}, CPMpy
examples~\cite{Guns2019}, models by
Kjellerstrand~\cite{Kjellerstrand2024}, and problems from the APLAI
course at KU Leuven. The collection covers diverse combinatorial
problem types, including scheduling, assignment, sequencing, graph
problems, and combinatorial design. With 30 optimization and 71
decision problems, the collection provides a balanced coverage of
constraint programming challenges.

Each problem in CP-Bench consists of a natural-language description in
Markdown format, input data, required output variables, and a
ground-truth model for verification. Solutions must output results in JSON format using the specified variable names. The
ground-truth models enable semantic validation through constraint satisfaction checking rather than syntactic correctness or execution success. The ground-truth models are withheld from the agent and used solely for post-hoc validation.

\subsection{Fixed Workflow Results}

Michailidis et al.~\cite{Michailidis2025} evaluated fixed workflow
approaches on CP-Bench, testing progressively detailed prompts
including basic instructions, modeling guidelines, and API
documentation, as well as retrieval-augmented in-context
learning. Their results indicate that combining repeated sampling with
self-verification using GPT-4-turbo achieves the highest accuracy of
70\%. A general limit of these fixed workflow approaches is that they
do not support testing hypotheses, debugging errors, or verifying partial solutions.

\subsection{Benchmark Corrections and Clarifications}

During testing, we encountered issues with some CP-Bench
problems. The CP-Bench authors acknowledged these issues
independently\footnote{personal communication, August 2025}, including ground-truth models with symmetry-breaking
constraints that reject valid solutions, unclear output format
requirements, and ambiguities in problem statements. Without a
clear problem specification, the modeling
ability cannot be evaluated separately from format guessing.
We therefore updated the CP-Bench problems to address these issues.

Our benchmark updates fall into three categories.

First, we clarified 31 ambiguous problem statements. For instance,
Problem 080 (TSP) required specifying whether Euclidean distances
should be rounded or truncated, which can lead to different distance
matrices and thus different optimal solutions. For Problem 013 (magic
square), we corrected a misleading description that stated the goal is
to find a ``magic square'' when the problem actually requires a
``magic sequence.'' Problem 060 (crossword) provided a visual grid
and word list, but required explicit specification of slot lengths and
intersection constraints to be solvable.

Second, we added structured JSON schemas to all problems to clarify
the required output format. We recommend this practice for any future
extensions of the benchmark collection, as it eliminates trivial but
annoying discrepancies, such as whether array indices are
supposed to start with 0 or 1.

Third, we corrected errors in 19 ground-truth models: we removed symmetry-breaking constraints from 10 and fixed logic errors in 9 others.

While these updates do not make the problems easier, they remove
confounding factors that obscure actual modeling capability. The
underlying combinatorial structure and search space complexity remain
the same; no changes hint at constraint formulations or modeling
strategies. The success rates reported in the original CP-Bench paper
were obtained without using our clarifications. A direct comparison
between our results and theirs is not possible, as the fixed workflow
can also benefit from the clarifications. The clarified
benchmark is available online~\cite{CPAgentData2025}.

\section{Experiments}

\subsection{Experimental Setup}
\label{sec:expsetup}

We ran all our experiments on all 101 CP-Bench problems three times,
with CP-Agent implemented as agentic-python-coder (version 2.2.1)
using Claude Sonnet 4.5 (version 20250929) as the language model for
our main experiment. The models used in the model generalization and
the task management experiments are described below. For each
problem, the agent produced a solution file containing the generated
CPMpy model, and we recorded a conversation log with the complete
agent interaction. We tested all solutions against the ground-truth
CPMpy models. For each satisfaction problem, we tested whether the
solution satisfies all ground-truth constraints. For optimization
problems, we compared the objective value of the generated solution
with an optimal pre-computed ground-truth value to confirm
optimality. Note that a problem may have several correct and optimal
solutions, hence the ground-truth model must not contain symmetry-breaking constraints. We always report results averaged across three
runs to account for randomness.

The agentic-python-coder, the problem specifications and ground-truth
models to reproduce our experiments, and the generated solutions and
execution logs, are available online~\cite{CPAgentData2025}. The
data is encrypted (password \texttt{Ckc1ADsKXRfmzD2}) to avoid
training data contamination.

\subsection{Results}
\label{sec:results}

Our experimental results show that CP-Agent achieves 100\% accuracy,
successfully solving all 101 CP-Bench problems across all three
independent runs. We validated all generated solutions against the
ground-truth models and confirmed that all solutions were correct
and optimal.

Our results are summarized in the table on the next page:
tool usage, token consumption, and execution time for all 101
CP-Bench problems (averaged over three independent runs).
Column \textbf{ex} shows the number of \texttt{python\_exec} calls (iterative refinement cycles).
Columns \textbf{in/out} show input/output tokens in thousands.
Column \textbf{time} shows execution time in seconds (rounded).

The number of \texttt{python\_exec} calls varies and depends on the structure of the problem, ranging from 1 to 28, with simpler problems requiring fewer iterations and complex scheduling or optimization problems requiring more exploration. Problem 014 (crossfig) stands out with an unusually large number of Python executions (28 average), followed by 009 (basketball) with 24 executions, reflecting their combinatorial complexity.

\begin{table}[htbp]
\centering
\label{tab:results}
\smallskip
\setlength{\tabcolsep}{4pt}
\begin{tabular}{@{}lrrrrllrrrr@{}}
\toprule
\textbf{Problem} & \textbf{ex} & \textbf{in} & \textbf{out} & \textbf{time} &
~~~~~~~&
\textbf{Problem} & \textbf{ex} & \textbf{in} & \textbf{out} & \textbf{time} \\
\cmidrule(r){1-5} \cmidrule(l){7-11}
001\_car\_seq     &  2.7 &   24 &   2 &   43 && 052\_circular\_t  &  3.7 &   29 &   2 &   47 \\
002\_template    &    2 &   21 &   2 &   52 && 053\_clock\_trip  &    1 &   14 &   1 &   26 \\
003\_autocorr    &  3.3 &   26 &   2 &   44 && 054\_cmo\_2012    &    3 &   25 &   3 &   49 \\
004\_golomb      &  2.3 &   22 &   2 &   42 && 055\_coin3\_appl  &    2 &   18 &   2 &   34 \\
005\_all\_inter   &  1.7 &   19 &   2 &   32 && 056\_coins\_grid  &  4.3 &   38 &   2 &   47 \\
006\_vessel      &  2.3 &   26 &   3 &   51 && 057\_contractin  &    1 &   14 &   1 &   25 \\
007\_perfect\_sq  &    4 &   30 &   2 &   44 && 058\_covering\_o  &    2 &   24 &   2 &   42 \\
008\_social\_gol  &  7.7 &   54 &   3 &   61 && 059\_crew        &  2.3 &   36 &   4 &   62 \\
009\_basketball  & 23.7 &  256 &  12 &  245 && 060\_crossword   &    4 &   40 &   3 &   55 \\
010\_nonogram    &    8 &   90 &   6 &  110 && 061\_crypta      &  4.7 &   67 &   4 &   74 \\
011\_progressive &    8 &   79 &   5 &   93 && 062\_eighteen\_h  &    1 &   12 &   1 &   21 \\
012\_schurs      &  2.7 &   23 &   2 &   36 && 063\_facility\_l  &    3 &   29 &   3 &   57 \\
013\_magic\_sq    &    1 &   15 &   1 &   22 && 064\_fifty\_puzz  &  1.3 &   15 &   1 &   23 \\
014\_crossfig    & 28.3 &  737 &  33 &  542 && 065\_three\_coin  &    1 &   15 &   1 &   27 \\
015\_langford    &  1.7 &   19 &   2 &   36 && 066\_three\_sum   &    1 &   13 &   1 &   20 \\
016\_sports      &  8.3 &   75 &   5 &   99 && 067\_twelve\_pac  &  1.3 &   14 &   1 &   25 \\
017\_bibd        &  2.7 &   24 &   2 &   37 && 068\_bus\_schedu  &    1 &   14 &   1 &   27 \\
018\_word\_des    &  6.3 &   47 &   3 &   80 && 069\_jobshop     &  3.3 &   29 &   3 &   52 \\
019\_steiner     &  3.7 &   29 &   2 &   41 && 070\_knapsack    &    1 &   13 &   1 &   22 \\
020\_num\_part    &  2.7 &   21 &   2 &   60 && 071\_mario       &  7.3 &   86 &   6 &  110 \\
021\_diamond     &  4.3 &   41 &   4 &   77 && 072\_minesweepe  &    5 &   37 &   3 &   57 \\
022\_graceful    &    1 &   16 &   1 &   26 && 073\_n\_puzzle    & 10.7 &  106 &   6 &  119 \\
023\_n\_queens    &    2 &   20 &   2 &   42 && 074\_packing\_re  &  8.3 &   75 &   5 &   83 \\
024\_costas      &    2 &   20 &   1 &   30 && 075\_resource\_c  &  5.7 &   68 &   4 &   78 \\
025\_hadamard    &  4.3 &   31 &   2 &   46 && 076\_room\_assig  &    2 &   21 &   2 &   38 \\
026\_abbots\_puz  &    1 &   13 &   1 &   23 && 077\_send\_more   &    1 &   14 &   1 &   25 \\
027\_added\_corn  &    2 &   20 &   2 &   37 && 078\_set\_game    &  3.3 &   34 &   3 &   60 \\
028\_ages\_of\_th  &    4 &   27 &   2 &   53 && 079\_sudoku      &    2 &   23 &   2 &   39 \\
029\_age\_changi  &  4.7 &   38 &   3 &   57 && 080\_tsp         &  3.3 &   24 &   2 &   40 \\
030\_allergy     &    2 &   21 &   2 &   44 && 081\_who\_killed  &    2 &   21 &   2 &   42 \\
031\_among       &    2 &   17 &   1 &   42 && 082\_wolf\_goat   &  3.3 &   28 &   3 &   58 \\
032\_appointmen  &    1 &   14 &   1 &   36 && 083\_zebra       &  3.7 &   38 &   4 &   59 \\
033\_archery\_pu  &  2.3 &   18 &   1 &   49 && 084\_bank\_card   &    1 &   13 &   1 &   23 \\
034\_arch\_frien  &    2 &   24 &   3 &   56 && 085\_climbing\_s  &  1.3 &   16 &   1 &   31 \\
035\_assignment  &  2.7 &   22 &   2 &   49 && 086\_color\_simp  &    2 &   18 &   1 &   30 \\
036\_autoref     &  4.3 &   31 &   2 &   46 && 087\_exodus      &    5 &   45 &   4 &   71 \\
037\_bales\_of\_h  &    4 &   33 &   3 &   66 && 088\_farmer\_and  &  2.3 &   19 &   2 &   34 \\
038\_bananas     &  1.7 &   18 &   2 &   39 && 089\_five\_floor  &    1 &   14 &   1 &   24 \\
039\_best\_host   &  1.7 &   18 &   2 &   33 && 090\_grocery     &  4.3 &   31 &   2 &   51 \\
040\_big\_bang2   &  1.7 &   21 &   2 &   39 && 091\_guards\_and  &    2 &   20 &   2 &   37 \\
041\_bin\_packin  &    1 &   13 &   1 &   23 && 092\_hardy\_1729  &  1.7 &   15 &   1 &   28 \\
042\_birthday\_c  &    1 &   14 &   1 &   26 && 093\_kidney\_exc  &  3.3 &   27 &   2 &   44 \\
043\_bowls\_and   &  1.7 &   16 &   1 &   30 && 094\_magic\_sq    &  1.3 &   15 &   1 &   28 \\
044\_broken\_wei  &    2 &   19 &   2 &   40 && 095\_maximal\_in  &  1.7 &   19 &   2 &   36 \\
045\_building\_b  &  2.3 &   20 &   2 &   34 && 096\_money\_chan  &    1 &   13 &   1 &   24 \\
046\_cabling     &  1.3 &   17 &   2 &   36 && 097\_movie\_sche  &  2.7 &   23 &   2 &   41 \\
047\_calvin\_puz  &  2.7 &   24 &   2 &   49 && 098\_people\_in   &    1 &   15 &   2 &   36 \\
048\_candies     &  1.3 &   16 &   2 &   28 && 099\_subsets\_10  &    1 &   14 &   1 &   27 \\
049\_capital\_bu  &    1 &   14 &   1 &   25 && 100\_subset\_sum  &  2.7 &   21 &   1 &   37 \\
050\_chess\_set   &  6.3 &   79 &   3 &   59 && 101\_thick\_as\_t  &  2.7 &   26 &   3 &   54 \\
051\_circling\_s  &  2.3 &   25 &   3 &   47 &&                 &      &      &     &      \\
\bottomrule
\end{tabular}
\end{table}
 
The table also shows the token consumption. On average we have 37k
input and 3k output tokens per problem. As expected, Problem 014
(crossfig), the problem with the largest number of exec calls has also
the largest token consumption (737k input tokens).

The specifics of execution times differ from problem to problem,
ranging from 20 seconds to 542 seconds, with an average of 52 seconds
across the three runs. The average run time per problem in each of the
three runs is 48, 57, and 52 seconds, respectively, which demonstrates
the stochastic nature of LLM-based systems, but can also reflect
variety in API response time and network speed.

\bigskip
We present representative runs on two problems to illustrate how the
agent approaches constraint modeling, with different problem-solving patterns.

\subsection{Problem 061: Cryptarithmetic Puzzle}

The problem is to assign letters with unique digits (0--9) such that
the given 20-digit addition equation holds. The log reveals how the
agent adapts when an initial approach proves unworkable.

In the first four executions, the agent introduces integer variables (one
per letter), an AllDifferent constraint, mappings from variables to
their positions in the three numbers, and constraints forcing leading
digits to be non-zero. This completes the description of the problem
structure.

The fifth execution attempts to treat each 20-digit number as a single
large integer variable. The agent constructs expressions like
\texttt{num1 = letter[0]*10**19 + letter[1]*10**18 + ...} for all
three numbers and adds the constraint \texttt{num1 + num2 ==
 result}. This approach fails because variable domains
of size $10^{20}$ exceed the solver's integer representation capacity, causing a \texttt{TypeError: \_\_init\_\_(): incompatible constructor arguments}.

With this feedback, the agent redesigns the model, introducing binary carry variables and models addition column by column, thereby mimicking manual arithmetic:

\begin{lstlisting}[language=Python]
carries = intvar(0, 1, shape=21)
for pos in range(20):
  digit1 = letter_to_var[num1_chars[19-pos]]
  digit2 = letter_to_var[num2_chars[19-pos]]
  result_digit = letter_to_var[result_chars[19-pos]]
  carry_in = carries[pos]
  carry_out = carries[pos+1]
  model += (digit1 + digit2 + carry_in
       == result_digit + 10 * carry_out)
\end{lstlisting}

This way, all intermediate values are kept within standard integer bounds, enabling the solver to determine the correct digit assignment.

\subsection{Problem 082: Wolf, Goat, and Cabbage}

The above example demonstrated error recovery; this well-known river-crossing
puzzle illustrates the agent's behavior when proceeding without failures.

The first step is to establish state variables: \texttt{wolf\_pos[t]},
\texttt{goat\_pos[t]}, \texttt{cabbage\_pos[t]}, and
\texttt{boat\_pos[t]}, which track locations (shore 0 or 1) across an
8-step timeline. The second step of the construction fixes boundary
conditions---all entities start at shore~0 and end at shore~1. The
third step encodes the movement mechanics: the boat crosses at each
step, the farmer transports at most one item per trip, and item
movements coincide with boat movements. The fourth step adds safety constraints:

\begin{lstlisting}[language=Python]
for i in range(steps):
  model += ~((goat_pos[i] == cabbage_pos[i])
        & (goat_pos[i] != boat_pos[i]))
  model += ~((wolf_pos[i] == goat_pos[i])
        & (wolf_pos[i] != boat_pos[i]))
\end{lstlisting}

These constraints ensure that the goat and cabbage (or wolf and goat) do not share a shore without the farmer present. The systematic layering---state, boundaries, movement, and safety---naturally reflects the construction process. Each phase is traversed in order to verify correctness before proceeding to the next. The execution completes successfully on the first attempt.

These two behaviors of the agent can be observed in different phases of problem solving: complete redesigns when facing fundamental obstacles and methodical incremental construction when the path is clear. The agent uses adaptive approaches rather than fixed workflows.

\section{Model Generalization}

To study whether the approach generalizes beyond Claude Sonnet 4.5, we
evaluated six frontier models on ten representative problems, with three
runs per model. The models come from four model makers: Anthropic
(Sonnet 4.5, Opus 4.5, Haiku 4.5), OpenAI (GPT-5.1-codex-mini), xAI
(Grok-code-fast-1), and DeepSeek (v3.2). Due to infrastructure issues
during the experimental period, we did not test with Google's Gemini
models. We ran each problem with a 20-minute timeout, using a
temperature of 0.0 for each model.

We give the results in Table~\ref{tab:multimodel}. The tested models
achieved a success rate of 80--100\%. Sonnet 4.5 maintained perfect
accuracy. Opus 4.5 and GPT-5.1 both reached 97\%. Haiku 4.5, Grok, and
DeepSeek achieved 80--87\% accuracy; most failures occurred on Problem 009 (basketball), though sporadic failures also affected other problems. Basketball requires coordinating nine constraint categories, including mirroring schemes, home/away patterns, rival pairings, and date-specific requirements---the highest constraint density in this subset.

For some runs, the solver timed out after 20 minutes. Looking at execution logs, we see that these agents were not converging toward correct solutions, suggesting the timeout did not cut short otherwise successful attempts.

Our results show that the architecture generalizes across model
families. This indicates that our agentic workflow generalizes from
our default model, Sonnet. Opus, considered superior to Sonnet, showed
slightly weaker performance, possibly because the prompt was
optimized for Sonnet. Prompt optimization frameworks such as DSPy~\cite{Khattab2024}
could be used to automatically tune the project prompt for specific models,
potentially narrowing the performance gap observed with other models.

\begin{table}[t]
\centering
\caption{Multi-model evaluation (3 runs each). Numbers indicate successful runs out of 3. Model success rates: Sonnet 100\%, Opus/GPT-5 97\%, Haiku 87\%, DeepSeek 83\%, Grok 80\%.}
\label{tab:multimodel}
\av{\smallskip}
\setlength{\tabcolsep}{5pt}
\begin{tabular}{@{}lcccccc@{}}
\toprule
\textbf{Problem} & \rotatebox{90}{Sonnet 4.5} & \rotatebox{90}{Opus 4.5} & \rotatebox{90}{Haiku 4.5} & \rotatebox{90}{GPT-5.1} & \rotatebox{90}{Grok} & \rotatebox{90}{DeepSeek} \\
\midrule
\texttt{001\_car\_seq} & 3 & 3 & 3 & 3 & 3 & 3 \\
\texttt{002\_template} & 3 & 2 & 3 & 3 & 3 & 2 \\
\texttt{004\_golomb} & 3 & 3 & 3 & 3 & 2 & 3 \\
\texttt{007\_perfect\_sq} & 3 & 3 & 3 & 3 & 3 & 3 \\
\texttt{009\_basketball} & 3 & 3 & 1 & 2 & 0 & 0 \\
\texttt{010\_nonogram} & 3 & 3 & 3 & 3 & 3 & 2 \\
\texttt{013\_magic\_sq} & 3 & 3 & 3 & 3 & 3 & 3 \\
\texttt{015\_langford} & 3 & 3 & 1 & 3 & 2 & 3 \\
\texttt{020\_num\_part} & 3 & 3 & 3 & 3 & 3 & 3 \\
\texttt{037\_bales\_of\_h} & 3 & 3 & 3 & 3 & 2 & 3 \\
\bottomrule
\end{tabular}
\end{table}

\section{Task Management Tool}

Explicit task tracking for LLM agents---maintaining a mutable list of
pending, in-progress, and completed tasks---was introduced by
BabyAGI~\cite{Nakajima2023BabyAGI} and formalized in plan-and-solve
prompting~\cite{Wang2023PlanSolve}. Our implementation provides a tool that realizes this pattern. Our architecture requires only \texttt{todo\_write}, not a corresponding \texttt{todo\_read}: without history compaction, all previous tool calls remain in the conversation context, so the agent can always see its earlier task updates. To evaluate whether explicit task management benefits constraint modeling, we compare our method with and without this tool.

For this experiment, we use a smaller model (Claude Haiku 4.5) than our default (Claude Sonnet 4.5) to reveal potential differences. We consider the following 40 problems, stratified by difficulty: 20 easy problems (solved in 27 seconds or less by Sonnet 4.5) and 20 hard problems (requiring 58 seconds or more). We run each problem three times with \texttt{todo\_write} disabled and three times with it enabled. The results are summarized in Table~\ref{tab:todo-comparison}.

\begin{table*}[t]
\centering
\caption{Task management ablation on 40 problems (Claude Haiku 4.5, 3 runs per condition). Columns: p~=~success (out of 3), tok~=~tokens (K, averaged), t~=~time (s, averaged). Bold highlights problems where success changed.}
\label{tab:todo-comparison}
\av{\smallskip}
\setlength{\tabcolsep}{3pt}
\begin{tabular}{@{}lrrr rrr c@{\qquad} lrrr rrr@{}}
\toprule
& \multicolumn{3}{c}{\textbf{no todo}} & \multicolumn{3}{c}{\textbf{todo}} && & \multicolumn{3}{c}{\textbf{no todo}} & \multicolumn{3}{c}{\textbf{todo}} \\
\cmidrule(r){2-4} \cmidrule(r){5-7} \cmidrule(r){10-12} \cmidrule(l){13-15}
\textbf{Easy Problem} & \textbf{p} & \textbf{tok} & \textbf{t} & \textbf{p} & \textbf{tok} & \textbf{t} && \textbf{Hard Problem} & \textbf{p} & \textbf{tok} & \textbf{t} & \textbf{p} & \textbf{tok} & \textbf{t} \\
\midrule
013\_magic\_sq  & 3 & 15 & 26 & 3 & 39 & 35 && 008\_social\_gol & 3 & 36 & 36 & 3 & 122 & 74 \\
022\_graceful  & 3 & 38 & 32 & 3 & 49 & 42 && \textbf{009\_basketball} & \textbf{3} & 752 & 303 & \textbf{0} & 181 & 119 \\
026\_abbots\_puz & 3 & 35 & 29 & 3 & 57 & 40 && 010\_nonogram   & 3 & 47 & 48 & 3 & 112 & 67 \\
041\_bin\_packin & 3 & 10 & 13 & 3 & 52 & 48 && \textbf{011\_progressive} & \textbf{1} & 201 & 111 & \textbf{2} & 730 & 197 \\
042\_birthday\_c & 3 & 19 & 20 & 3 & 52 & 40 && \textbf{014\_crossfig}  & \textbf{0} &1986 & 617 & \textbf{2} &1592 & 492 \\
049\_capital\_bu & 3 & 10 & 18 & 3 & 37 & 41 && \textbf{016\_sports}   & \textbf{0} & 112 & 70 & \textbf{1} & 712 & 228 \\
\textbf{053\_clock\_trip} & \textbf{2} & 17 & 18 & \textbf{3} & 87 & 54 && 018\_word\_des  & 3 & 41 & 40 & 3 & 160 & 90 \\
057\_contractin & 3 & 16 & 18 & 3 & 38 & 35 && 021\_diamond   & 3 & 69 & 63 & 3 & 130 & 77 \\
062\_eighteen\_h & 3 & 15 & 16 & 3 & 57 & 49 && 037\_bales\_of\_h & 3 & 40 & 40 & 3 & 149 & 81 \\
064\_fifty\_puzz & 3 & 10 & 12 & 3 & 35 & 29 && 050\_chess\_set  & 3 & 10 & 16 & 3 & 38 & 31 \\
065\_three\_coin & 3 & 15 & 16 & 3 & 70 & 60 && \textbf{059\_crew}    & \textbf{3} & 80 & 55 & \textbf{0} & 137 & 75 \\
066\_three\_sum & 3 &  9 & 13 & 3 & 31 & 26 && \textbf{061\_crypta}   & \textbf{3} & 43 & 31 & \textbf{2} & 297 & 127 \\
067\_twelve\_pac & 3 &  9 & 12 & 3 & 39 & 33 && \textbf{071\_mario}    & \textbf{3} & 243 & 145 & \textbf{2} & 291 & 135 \\
068\_bus\_schedu & 3 & 18 & 20 & 3 & 50 & 39 && 073\_n\_puzzle  & 3 & 203 & 95 & 3 & 273 & 122 \\
070\_knapsack  & 3 &  9 & 11 & 3 & 37 & 32 && 074\_packing\_re & 3 & 22 & 25 & 3 & 134 & 81 \\
077\_send\_more & 3 & 11 & 13 & 3 & 41 & 37 && 075\_resource\_c & 3 & 55 & 51 & 3 & 86 & 61 \\
084\_bank\_card & 3 & 10 & 16 & 3 & 26 & 24 && 078\_set\_game  & 3 & 24 & 24 & 3 & 105 & 70 \\
089\_five\_floor & 3 & 11 & 14 & 3 & 33 & 30 && \textbf{082\_wolf\_goat} & \textbf{3} & 62 & 60 & \textbf{0} & 286 & 156 \\
096\_money\_chan & 3 & 13 & 22 & 3 & 32 & 34 && 083\_zebra    & 3 & 30 & 32 & 3 & 150 & 91 \\
099\_subsets\_10 & 3 & 16 & 19 & 3 & 28 & 26 && 087\_exodus    & 3 & 123 & 88 & 3 & 223 & 106 \\
\midrule
\textbf{Easy (60 runs)} & 59 & & & 60 & & && \textbf{Hard (60 runs)} & 52 & & & 45 & & \\
\bottomrule
\end{tabular}
\end{table*}

On easy problems, compared with the baseline, \texttt{todo\_write}
results in a significant increase in token consumption (191\%, from
15K to 44K average) and execution time (111\%, from 18s to 38s), while
improving success by just one run. On hard problems,
\texttt{todo\_write} even reduced success from 52/60 to 45/60 while
still adding overhead in terms of token usage and longer execution
time.

A closer look at individual problems reveals a pattern. The task
management tool provides structure for extended problem-solving, which
helps three previously failing hard problems---crossfig, sports, and
progressive. In contrast, it causes failures on three problems that worked without
it. The task-tracking overhead appears to distract the agent from finding solutions in some cases.

\section{MCP Server}

The Model Context Protocol (MCP) is a standardized interface for connecting language models with external tools and data sources~\cite{Hou2025MCP}. MCP was initially introduced by Anthropic in late 2024~\cite{MCP2024} as a tool for agent-tool interaction, and has since been adopted by major AI platforms and integrated into development environments. In December 2025, MCP governance transferred to the Linux Foundation.

To make our framework accessible from other agentic systems, beyond
our own ReAct agent, we define the same persistent IPython kernel
system as an MCP server (\texttt{ipython\_mcp}). This allows it to be
integrated with any MCP-compatible client. The MCP server is equipped
with \texttt{python\_exec} for code execution, the same tool central
to our standalone agent. It also comprises three tools for controlling
the coding infrastructure: \texttt{python\_reset} to clear state and
install packages, \texttt{python\_status} to query session
information, and \texttt{python\_interrupt} to halt long-running
computations. This interactive mode also serves an educational
purpose: users can observe how natural language constraints are
formalized into CPMpy models, request explanations, and refine models
based on solver feedback. Once developed, the generated code can be
deployed to other contexts.

This approach complements the MCP-Solver~\cite{Szeider2025}, which
provides specialized tools for constraint modeling (MiniZinc, PySAT,
Z3, ASP). In contrast to the MCP-Solver, our new \texttt{ipython\_mcp} server provides general-purpose Python execution, with domain adaptation driven by prompts rather than tool specialization.

\section{Conclusion}

Our experiments provide several insights into the design and
evaluation of agentic systems for constraint modeling.

Our results show that agentic workflows with execution feedback can reliably automate constraint modeling from natural language.

Our main experiment demonstrates that a short prompt performs very well (even
outperforms an 800-line earlier version). This suggests that, for
foundation models with constraint-programming knowledge, the role of
the prompt is to set high-level strategy and the output format rather
than to provide exhaustive procedural instructions. This is in line with
recent results on prompt compression showing that shorter, denser
prompts can match or exceed verbose
prompts~\cite{Jiang2024LongLLMLingua}. In terms of procedural detail,
excessive instruction might reduce the agent's ability to solve
problems effectively.

Our model generalization experiments show that our agentic architecture is robust: six models from four providers achieved success rates of 80--100\%, with performance correlating with model capability.

Our experiment with the task management tool gives a mixed picture. While
the tool reduces the overall success rate and significantly increases time
and token usage, it was helpful for a few problems. Based on this outcome,
we turn off the tool by default but keep it in the codebase for further
testing, as there might be problem families where it shows a positive
overall performance.

Finally, we reported on further integration possibilities of our
system via the Model Context Protocol.

We believe that separating domain expertise from execution infrastructure, as
realized in our system, can be beneficial in other scientific computing
domains. In follow-up work, we used our system to transform natural
language descriptions into logic programs~\cite{Szeider2025ASPBench}, using
the same architecture with only a domain-specific project prompt.


\begin{thebibliography}{10}

\bibitem{MCP2024}
{Anthropic}.
\newblock Model {Context} {Protocol}, 2024.
\newblock Protocol specification for connecting language models to external
  systems.
\newblock URL: \url{https://modelcontextprotocol.io}.

\bibitem{uv2025}
{Astral Software Inc.}
\newblock uv: An extremely fast {Python} package and project manager, 2025.
\newblock Documentation: \url{https://docs.astral.sh/uv/}.
\newblock URL: \url{https://github.com/astral-sh/uv}.

\bibitem{DBLP:conf/icse/BouzeniaDP25}
Islem Bouzenia, Premkumar~T. Devanbu, and Michael Pradel.
\newblock Repairagent: An autonomous, llm-based agent for program repair.
\newblock In {\em 47th {IEEE/ACM} International Conference on Software
  Engineering, {ICSE} 2025, Ottawa, ON, Canada, April 26 - May 6, 2025}, pages
  2188--2200. {IEEE}, 2025.
\newblock \href {https://doi.org/10.1109/ICSE55347.2025.00157}
  {\path{doi:10.1109/ICSE55347.2025.00157}}.

\bibitem{ICSE24:ClassEval}
Xueying Du, Mingwei Liu, Kaixin Wang, Hanlin Wang, Junwei Liu, Yixuan Chen,
  Jiayi Feng, Chaofeng Sha, Xin Peng, and Yiling Lou.
\newblock Evaluating large language models in class-level code generation.
\newblock In {\em Proceedings of the IEEE/ACM 46th International Conference on
  Software Engineering}, ICSE '24, New York, NY, USA, 2024. Association for
  Computing Machinery.
\newblock \href {https://doi.org/10.1145/3597503.3639219}
  {\path{doi:10.1145/3597503.3639219}}.

\bibitem{Freuder2014}
Eugene~C. Freuder and Barry O'Sullivan.
\newblock Grand challenges for constraint programming.
\newblock {\em Constraints An Int. J.}, 19(2):150--162, 2014.
\newblock \href {https://doi.org/10.1007/S10601-013-9155-1}
  {\path{doi:10.1007/S10601-013-9155-1}}.

\bibitem{Gent1999}
Ian~P. Gent and Toby Walsh.
\newblock Csp\({}_{\mbox{lib}}\): {A} benchmark library for constraints.
\newblock In Joxan Jaffar, editor, {\em Principles and Practice of Constraint
  Programming - CP'99, 5th International Conference, Alexandria, Virginia, USA,
  October 11-14, 1999, Proceedings}, volume 1713 of {\em Lecture Notes in
  Computer Science}, pages 480--481. Springer, 1999.
\newblock \href {https://doi.org/10.1007/978-3-540-48085-3\_36}
  {\path{doi:10.1007/978-3-540-48085-3\_36}}.

\bibitem{Guns2019}
Tias Guns.
\newblock Modeling and solving with {CPMpy}, 2025.
\newblock {CPMpy} documentation.
\newblock URL: \url{https://cpmpy.readthedocs.io/}.

\bibitem{Hou2025MCP}
Xinyi Hou, Yanjie Zhao, Shenao Wang, and Haoyu Wang.
\newblock Model context protocol ({MCP}): Landscape, security threats, and
  future research directions.
\newblock {\em CoRR}, abs/2503.23278, 2025.
\newblock \href {https://arxiv.org/abs/2503.23278} {\path{arXiv:2503.23278}},
  \href {https://doi.org/10.48550/arXiv.2503.23278}
  {\path{doi:10.48550/arXiv.2503.23278}}.

\bibitem{Jiang2024LongLLMLingua}
Huiqiang Jiang, Qianhui Wu, Xufang Luo, Dongsheng Li, Chin-Yew Lin, Yuqing
  Yang, and Lili Qiu.
\newblock {LongLLMLingua}: Accelerating and enhancing {LLMs} in long context
  scenarios via prompt compression.
\newblock In {\em Proceedings of the 62nd Annual Meeting of the Association for
  Computational Linguistics (ACL 2024)}, pages 1658--1677, 2024.
\newblock URL: \url{https://aclanthology.org/2024.acl-long.91/}.

\bibitem{Khattab2024}
Omar Khattab, Arnav Singhvi, Paridhi Maheshwari, Zhiyuan Zhang, Keshav
  Santhanam, Sri Vardhamanan, Saiful Haq, Ashutosh Sharma, Thomas~T. Joshi,
  Hanna Moazam, Heather Miller, Matei Zaharia, and Christopher Potts.
\newblock {DSPy}: Compiling declarative language model calls into
  state-of-the-art pipelines.
\newblock In {\em International Conference on Learning Representations (ICLR)},
  2024.

\bibitem{Kjellerstrand2024}
Håkan Kjellerstrand.
\newblock Constraint programming models collection, 2024.
\newblock Collection of constraint programming models in various languages.
\newblock URL: \url{https://github.com/hakank/hakank}.

\bibitem{LangGraph2024}
{LangChain}.
\newblock {LangGraph}: Build resilient language agents as graphs, 2024.
\newblock Python library for building stateful, multi-actor applications with
  {LLMs}.
\newblock URL: \url{https://python.langchain.com/docs/langgraph}.

\bibitem{DBLP:conf/nips/LiuXWZ23}
Jiawei Liu, Chunqiu~Steven Xia, Yuyao Wang, and Lingming Zhang.
\newblock Is your code generated by chatgpt really correct? rigorous evaluation
  of large language models for code generation.
\newblock In Alice Oh, Tristan Naumann, Amir Globerson, Kate Saenko, Moritz
  Hardt, and Sergey Levine, editors, {\em Advances in Neural Information
  Processing Systems 36: Annual Conference on Neural Information Processing
  Systems 2023, NeurIPS 2023, New Orleans, LA, USA, December 10 - 16, 2023},
  2023.
\newblock URL:
  \url{http://papers.nips.cc/paper_files/paper/2023/hash/43e9d647ccd3e4b7b5baab53f0368686-Abstract-Conference.html}.

\bibitem{Michailidis2025}
Kostis Michailidis, Dimos Tsouros, and Tias Guns.
\newblock {CP-Bench}: Evaluating large language models for constraint
  modelling.
\newblock {\em CoRR}, abs/2506.06052, 2025.
\newblock \href {https://arxiv.org/abs/2506.06052} {\path{arXiv:2506.06052}},
  \href {https://doi.org/10.48550/ARXIV.2506.06052}
  {\path{doi:10.48550/ARXIV.2506.06052}}.

\bibitem{Nakajima2023BabyAGI}
Yohei Nakajima.
\newblock {BabyAGI}: Task-driven autonomous agent, 2023.
\newblock {GitHub} repository, March 2023.
\newblock URL: \url{https://github.com/yoheinakajima/babyagi}.

\bibitem{Perez2007}
Fernando P{\'{e}}rez and Brian~E. Granger.
\newblock {IPython}: {A} system for interactive scientific computing.
\newblock {\em Comput. Sci. Eng.}, 9(3):21--29, 2007.
\newblock \href {https://doi.org/10.1109/MCSE.2007.53}
  {\path{doi:10.1109/MCSE.2007.53}}.

\bibitem{Ramamonjison2022}
Rindranirina Ramamonjison, Timothy T.~L. Yu, Raymond Li, Haley Li, Giuseppe
  Carenini, Bissan Ghaddar, Shiqi He, Mahdi Mostajabdaveh, Amin
  Banitalebi{-}Dehkordi, Zirui Zhou, and Yong Zhang.
\newblock {NL4Opt} competition: Formulating optimization problems based on
  their natural language descriptions.
\newblock In Marco Ciccone, Gustavo Stolovitzky, and Jacob Albrecht, editors,
  {\em {NeurIPS} 2022 Competition Track, November 28 - December 9, 2022,
  Online}, volume 220 of {\em Proceedings of Machine Learning Research}, pages
  189--203. {PMLR}, 2022.
\newblock URL: \url{https://proceedings.mlr.press/v220/ramamonjison22a.html}.

\bibitem{Regin2024CPLLM}
Florian Régin, Elisabetta~De Maria, and Alexandre Bonlarron.
\newblock Combining constraint programming reasoning with large language model
  predictions.
\newblock In {\em 30th International Conference on Principles and Practice of
  Constraint Programming (CP 2024)}, volume 307 of {\em LIPIcs}, pages
  25:1--25:21, 2024.
\newblock \href {https://doi.org/10.4230/LIPIcs.CP.2024.25}
  {\path{doi:10.4230/LIPIcs.CP.2024.25}}.

\bibitem{Schick2023}
Timo Schick, Jane Dwivedi{-}Yu, Roberto Dess{\`{\i}}, Roberta Raileanu, Maria
  Lomeli, Eric Hambro, Luke Zettlemoyer, Nicola Cancedda, and Thomas Scialom.
\newblock Toolformer: Language models can teach themselves to use tools.
\newblock In Alice Oh, Tristan Naumann, Amir Globerson, Kate Saenko, Moritz
  Hardt, and Sergey Levine, editors, {\em Advances in Neural Information
  Processing Systems 36: Annual Conference on Neural Information Processing
  Systems 2023, NeurIPS 2023, New Orleans, LA, USA, December 10 - 16, 2023},
  2023.
\newblock URL:
  \url{http://papers.nips.cc/paper\_files/paper/2023/hash/d842425e4bf79ba039352da0f658a906-Abstract-Conference.html}.

\bibitem{AgenticCoder2025}
Stefan Szeider.
\newblock Agentic {Python} coder, 2025.
\newblock {GitHub} repository.
\newblock URL: \url{https://github.com/szeider/agentic-python-coder}.

\bibitem{Szeider2025}
Stefan Szeider.
\newblock {Bridging Language Models and Symbolic Solvers via the Model Context
  Protocol}.
\newblock In Jeremias Berg and Jakob Nordstr\"{o}m, editors, {\em 28th
  International Conference on Theory and Applications of Satisfiability Testing
  (SAT 2025)}, volume 341 of {\em Leibniz International Proceedings in
  Informatics (LIPIcs)}, pages 30:1--30:12, Dagstuhl, Germany, 2025. Schloss
  Dagstuhl -- Leibniz-Zentrum f{\"u}r Informatik.
\newblock URL:
  \url{https://drops.dagstuhl.de/entities/document/10.4230/LIPIcs.SAT.2025.30}.

\bibitem{CPAgentData2025}
Stefan Szeider.
\newblock Supplementary material for {CP-Agent}: Agentic constraint
  programming, 2025.
\newblock Zenodo.
\newblock \href {https://doi.org/10.5281/zenodo.18034815}
  {\path{doi:10.5281/zenodo.18034815}}.

\bibitem{Szeider2025ASPBench}
Stefan Szeider.
\newblock {ASP-Bench}: From natural language to logic programs.
\newblock In {\em 2nd International Workshop on Neuro-Symbolic Software
  Engineering (NSE)}, 2026.
\newblock To appear at ICSE 2026. Preprint at \url{https://arxiv.org/abs/2602.01171}.

\bibitem{Tsouros2023HolyGrail}
Dimos Tsouros, Hélène Verhaeghe, Serdar Kadıoğlu, and Tias Guns.
\newblock Holy grail 2.0: From natural language to constraint models.
\newblock {\em CoRR}, abs/2308.01589, 2023.
\newblock URL: \url{https://arxiv.org/abs/2308.01589}, \href
  {https://arxiv.org/abs/2308.01589} {\path{arXiv:2308.01589}}.

\bibitem{Tyagi2024}
Nemika Tyagi, Mihir Parmar, Mohith Kulkarni, Aswin RRV, Nisarg Patel, Mutsumi
  Nakamura, Arindam Mitra, and Chitta Baral.
\newblock Step-by-step reasoning to solve grid puzzles: Where do {LLMs} falter?
\newblock In Yaser Al{-}Onaizan, Mohit Bansal, and Yun{-}Nung Chen, editors,
  {\em Proceedings of the 2024 Conference on Empirical Methods in Natural
  Language Processing, {EMNLP} 2024, Miami, FL, USA, November 12-16, 2024},
  pages 19898--19915. Association for Computational Linguistics, 2024.
\newblock \href {https://doi.org/10.18653/V1/2024.EMNLP-MAIN.1111}
  {\path{doi:10.18653/V1/2024.EMNLP-MAIN.1111}}.

\bibitem{Wallace1996}
Mark Wallace.
\newblock Practical applications of constraint programming.
\newblock {\em Constraints An Int. J.}, 1(1/2):139--168, 1996.
\newblock \href {https://doi.org/10.1007/BF00143881}
  {\path{doi:10.1007/BF00143881}}.

\bibitem{Wang2023PlanSolve}
Lei Wang, Wanyu Xu, Yihuai Lan, Zhiqiang Hu, Yunshi Lan, Roy~Ka{-}Wei Lee, and
  Ee{-}Peng Lim.
\newblock Plan-and-solve prompting: Improving zero-shot chain-of-thought
  reasoning by large language models.
\newblock In {\em Proceedings of the 61st Annual Meeting of the Association for
  Computational Linguistics (ACL 2023)}, pages 2609--2634. Association for
  Computational Linguistics, 2023.
\newblock \href {https://doi.org/10.18653/V1/2023.ACL-LONG.147}
  {\path{doi:10.18653/V1/2023.ACL-LONG.147}}.

\bibitem{DBLP:conf/icml/WangCYZLPJ24}
Xingyao Wang, Yangyi Chen, Lifan Yuan, Yizhe Zhang, Yunzhu Li, Hao Peng, and
  Heng Ji.
\newblock Executable code actions elicit better {LLM} agents.
\newblock In {\em Forty-first International Conference on Machine Learning,
  {ICML} 2024, Vienna, Austria, July 21-27, 2024}. OpenReview.net, 2024.
\newblock URL: \url{https://openreview.net/forum?id=jJ9BoXAfFa}.

\bibitem{Wei2022}
Jason Wei, Xuezhi Wang, Dale Schuurmans, Maarten Bosma, Brian Ichter, Fei Xia,
  Ed~H. Chi, Quoc~V. Le, and Denny Zhou.
\newblock Chain-of-thought prompting elicits reasoning in large language
  models.
\newblock In Sanmi Koyejo, S.~Mohamed, A.~Agarwal, Danielle Belgrave, K.~Cho,
  and A.~Oh, editors, {\em Advances in Neural Information Processing Systems
  35: Annual Conference on Neural Information Processing Systems 2022, NeurIPS
  2022, New Orleans, LA, USA, November 28 - December 9, 2022}, 2022.
\newblock URL:
  \url{http://papers.nips.cc/paper\_files/paper/2022/hash/9d5609613524ecf4f15af0f7b31abca4-Abstract-Conference.html}.

\bibitem{DBLP:conf/issta/XiaZ24}
Chunqiu~Steven Xia and Lingming Zhang.
\newblock Automated program repair via conversation: Fixing 162 out of 337 bugs
  for {\textdollar}0.42 each using chatgpt.
\newblock In Maria Christakis and Michael Pradel, editors, {\em Proceedings of
  the 33rd {ACM} {SIGSOFT} International Symposium on Software Testing and
  Analysis, {ISSTA} 2024, Vienna, Austria, September 16-20, 2024}, pages
  819--831. {ACM}, 2024.
\newblock \href {https://doi.org/10.1145/3650212.3680323}
  {\path{doi:10.1145/3650212.3680323}}.

\bibitem{DBLP:conf/nips/YangJWLYNP24}
John Yang, Carlos~E. Jimenez, Alexander Wettig, Kilian Lieret, Shunyu Yao,
  Karthik Narasimhan, and Ofir Press.
\newblock Swe-agent: Agent-computer interfaces enable automated software
  engineering.
\newblock In Amir Globersons, Lester Mackey, Danielle Belgrave, Angela Fan,
  Ulrich Paquet, Jakub~M. Tomczak, and Cheng Zhang, editors, {\em Advances in
  Neural Information Processing Systems 38: Annual Conference on Neural
  Information Processing Systems 2024, NeurIPS 2024, Vancouver, BC, Canada,
  December 10 - 15, 2024}, 2024.
\newblock URL:
  \url{http://papers.nips.cc/paper_files/paper/2024/hash/5a7c947568c1b1328ccc5230172e1e7c-Abstract-Conference.html}.

\bibitem{Yao2023}
Shunyu Yao, Jeffrey Zhao, Dian Yu, Nan Du, Izhak Shafran, Karthik~R.
  Narasimhan, and Yuan Cao.
\newblock {ReAct}: Synergizing reasoning and acting in language models.
\newblock In {\em The Eleventh International Conference on Learning
  Representations, {ICLR} 2023, Kigali, Rwanda, May 1-5, 2023}. OpenReview.net,
  2023.
\newblock URL: \url{https://openreview.net/forum?id=WE\_vluYUL-X}.

\end{thebibliography}

\end{document}